\documentclass{article}

\usepackage[preprint]{aaj2026}
\usepackage[numbers]{natbib}

\usepackage[utf8]{inputenc}
\usepackage[T1]{fontenc}
\usepackage{hyperref}
\usepackage{url}
\usepackage{booktabs}
\usepackage{amsfonts}
\usepackage{amssymb}
\usepackage{nicefrac}
\usepackage{microtype}
\usepackage{xcolor}
\usepackage{graphicx}
\usepackage[section]{placeins} 
\usepackage{float} 
\usepackage{amsmath}
\usepackage{xspace}
\usepackage{pgfplots}
\pgfplotsset{compat=1.18}
\usepgfplotslibrary{polar}
\usepackage{tikz}
\usetikzlibrary{arrows.meta,positioning,calc}

\newcommand{\bench}{\textsc{TakeoffBench-v1}\xspace}
\newcommand{\anno}[1]{\tikz[baseline=-0.65ex]\node[circle,fill=blue!65!black,text=white,inner sep=1.05pt,font=\bfseries\tiny]{#1};}

\title{Handoff-H1: An Orchestrated Vision-Agent System for Material Quantity Takeoff from Construction Blueprints}

\author{%
  Bruno Chicelli \\
  Handoff AI Research \\
  bruno@handoff.ai \\
  \And
  Henrique Alves \\
  Handoff AI Research \\
  henrique@handoff.ai \\
  \And
  Rodrigo Anselmo \\
  Handoff AI Research \\
  rodrigo@handoff.ai \\
  \And
  Joshua Weinberg \\
  Handoff AI Research \\
  joshua@handoff.ai \\
  \And
  Felipe Lemos \\
  Handoff AI Research \\
  felipe@handoff.ai \\
  \And
  Jan Baryla \\
  Handoff AI Research \\
  jan@handoff.ai \\  
}

\begin{document}

\maketitle

\begin{abstract}
  Converting a set of architectural blueprints into a complete material quantity takeoff requires visual perception across drawing sheets, dimensional and multi-hop reasoning, and grounding in construction conventions that the drawings never state. We present \textbf{Handoff-H1}, a generally available takeoff system built from three layers that earn their results in combination: purpose-built computer-vision models that recover typed primitives from each sheet; \emph{vision agents}---tool-using agents equipped with generic image operations and in-house visual-task tools, including CV-model-backed counting and detection and the decomposition of framing plans into individual members; and a persistent, hierarchically structured project foundation, grounded in a curated construction knowledge base and consumed by orchestrated trade-scoped estimation with an independent verification pass. We evaluate on the Construction Blueprint Takeoff Benchmark (\bench): 10 real, permissioned, PII-stripped residential blueprint sets paired with consensus-validated expert takeoffs---2{,}009 verified line items, restricted for scoring to the 1{,}348 primary-tier materials that drive an estimate---scored per trade by an LLM judge on material \textbf{coverage} and quantity \textbf{Precision@25\%} (P@.25) and combined into a precision-weighted composite. Under identical scoring from the raw PDF, seven frontier and open-weight models span composites of 35--61, and independent professional estimators---scored against the same reconciled gold standard---post 77.6\% (65.5\% coverage, 87.9\% P@.25). Handoff-H1, working end-to-end from the raw PDF, reaches \textbf{81.6\%} (86.1\% coverage, 78.8\% P@.25): roughly 20 points above the strongest frontier agent, and above the independent estimators by pairing near-human quantity precision with coverage they do not reach. The evaluation harness is public for the open \texttt{harbor} framework; the blueprint sets and ground truth are available upon request for research use.
\end{abstract}

\section{Introduction}
\label{sec:intro}

\begin{figure}[t!]
  \centering
  \begin{tikzpicture}[
    font=\footnotesize,
    badge/.style={circle,fill=blue!65!black,text=white,inner sep=1.3pt,font=\bfseries\scriptsize,draw=white,line width=0.5pt},
  ]
    \node[anchor=south west,inner sep=0] (plan) {\includegraphics[height=7.6cm]{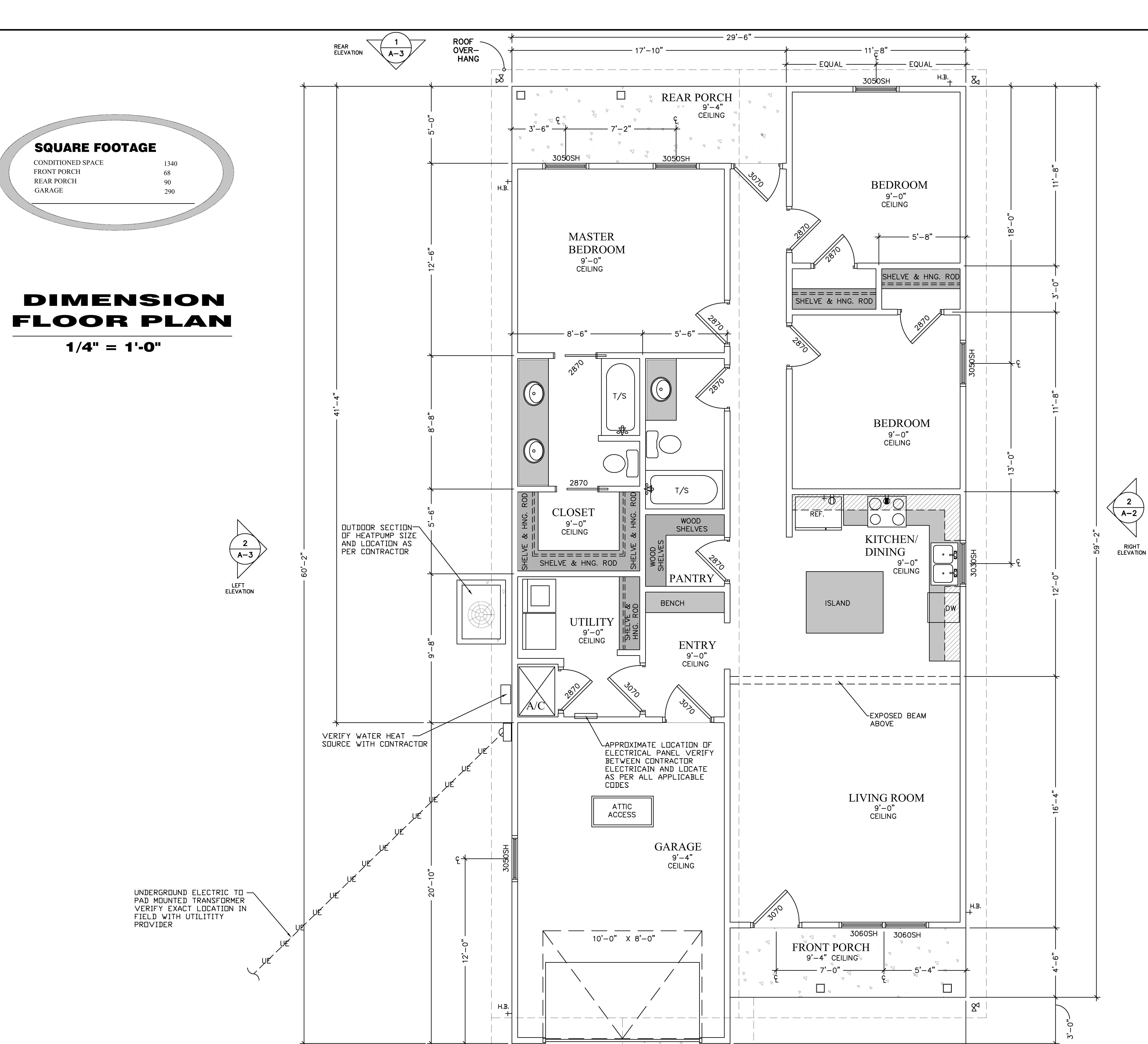}};
    \begin{scope}[x={($(plan.south east)-(plan.south west)$)},y={($(plan.north west)-(plan.south west)$)},shift={(plan.south west)}]
      \node[badge] at (0.676,0.960) {1};
      \node[badge] at (0.505,0.245) {2};
      \node[badge] at (0.408,0.700) {3};
    \end{scope}
    \node[anchor=north west,text width=4.55cm,align=left,inner sep=0] (panel) at ($(plan.north east)+(4.5mm,0)$) {%
      \textbf{What the sheet demands}\\[2.5pt]
      \anno{1}~\textbf{Measure.} Chained dimension strings at $\nicefrac{1}{4}''\!=\!1'$-$0''$; every wall length is arithmetic over the chains, read at sheet scale.\\[3.5pt]
      \anno{2}~\textbf{Reconcile.} The electrical panel exists only as a note (``verify with contractor''); notes, schedules, and plan must agree before anything is counted.\\[3.5pt]
      \anno{3}~\textbf{Apply convention.} Nothing on the sheet says the walls resolve into studs at 16$''$ on center in precut lengths---the takeoff depends on it anyway.\\[7pt]
      \textbf{What must come out} {\scriptsize(excerpt)}\\[2.5pt]
      {\scriptsize
      \begin{tabular}{@{}llr@{}}
        \toprule
        Trade & Material & Qty \\
        \midrule
        Framing & Exterior wall studs   & 264 EA \\
        Framing & Exterior top plates   & 528 LF \\
        Drywall & Ceilings, 1st floor   & 1{,}340 SF \\
        Roofing & Metal roof panels     & 2{,}235 SF \\
        \bottomrule
      \end{tabular}}%
    };
  \end{tikzpicture}
  \caption{The problem in one sheet. A dimension floor plan from a residential drawing set (title block removed), the three demands it makes of its reader, and four of the line items it must yield (quantities from Handoff-H1's takeoff of this set). A set like this one produces on the order of a hundred primary line items across nine trades, and none of the quantities on the right is written anywhere on the left.}
  \label{fig:anatomy}
\end{figure}

Construction cost estimation is a discipline in which a single error in material quantities can cascade into tens of thousands of dollars of overrun on a modest residential project. The task appears deceptively simple---count doors, measure walls, calculate roof area---but it requires (i) dimensional extraction across multiple drawing sheets, (ii) multi-hop reasoning that combines floor-plan dimensions with general-notes specifications and trade-level conventions, and (iii) grounding in domain knowledge that is rarely fully specified in the drawings themselves (Figure~\ref{fig:anatomy}). Framing lumber is spaced at 16 inches on center; drywall comes in 4$\times$8 sheets; shingle coverage depends on roof pitch. None of these conventions appears in the dimension lines. The scale compounds the difficulty: each drawing set in our benchmark yields between 69 and 166 primary-material line items across nine trades, and every one of them is the end point of that same read--reconcile--apply chain.

\begin{figure}[t]
  \centering
  \begin{tikzpicture}[font=\footnotesize]
    \node[anchor=south west,inner sep=0] (det) {\includegraphics[width=0.5\linewidth]{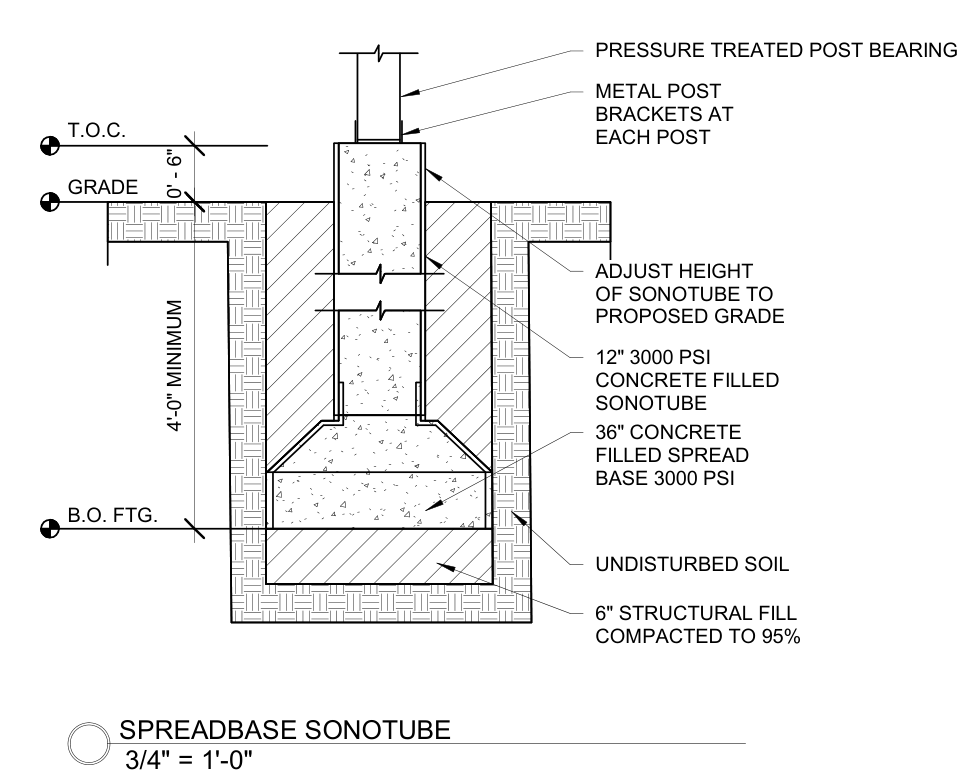}};
    \begin{scope}[x={($(det.south east)-(det.south west)$)},y={($(det.north west)-(det.south west)$)},shift={(det.south west)}]
      \coordinate (c-tube)   at (0.86,0.520);
      \coordinate (c-base)   at (0.86,0.415);
      \coordinate (c-fill)   at (0.86,0.195);
    \end{scope}
    \node[anchor=west,text width=4.7cm,align=left,inner sep=1pt] (f1) at ($(det.east)+(9mm,2.1cm)$)
      {\textbf{Concrete} --- 12$''$ tube forms (EA); 3000\,psi mix volume (CY)};
    \node[anchor=west,text width=4.7cm,align=left,inner sep=1pt] (f2) at ($(det.east)+(9mm,0.7cm)$)
      {\textbf{Concrete} --- 36$''$ spread bases join the same pour};
    \node[anchor=west,text width=4.7cm,align=left,inner sep=1pt] (f3) at ($(det.east)+(9mm,-0.7cm)$)
      {\textbf{Site work} --- 6$''$ compacted fill under every footing (CY)};
    \node[anchor=west,text width=4.7cm,align=left,inner sep=1pt] (f4) at ($(det.east)+(9mm,-2.1cm)$)
      {\textbf{Excavation} --- the 4$'$-$0''$ minimum depth on the left edge sets the dig volume (CY)};
    \draw[->,>={Stealth[length=1.6mm]},blue!65!black,thick] (c-tube) -- (f1.west);
    \draw[->,>={Stealth[length=1.6mm]},blue!65!black,thick] (c-base) -- (f2.west);
    \draw[->,>={Stealth[length=1.6mm]},blue!65!black,thick] (c-fill) -- (f3.west);
  \end{tikzpicture}
  \caption{A typical foundation detail---a spread-base sonotube footing, drawn at its own scale ($\nicefrac{3}{4}''=1'$-$0''$). A single callout block fans out into quantities across three scopes, and none of them appears as a number anywhere on the sheet. Reading the words is not enough; a takeoff must attach each one to the right trade and assembly.}
  \label{fig:detail}
\end{figure}
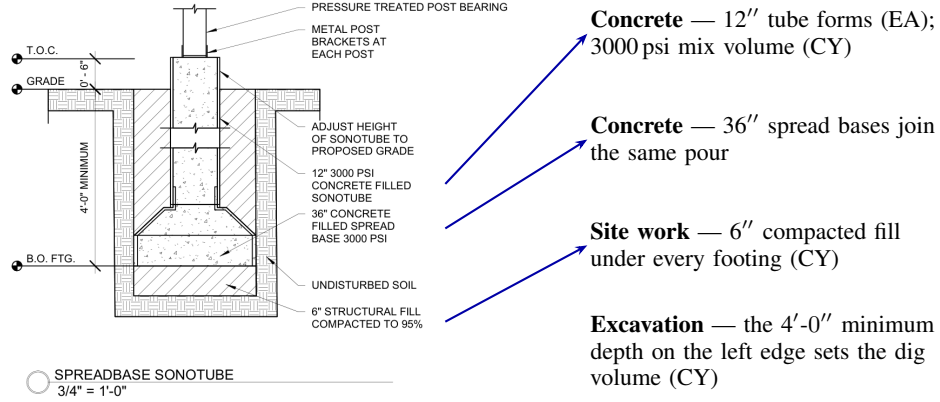

This combination defeats general-purpose systems from two directions at once. Large language models given raw blueprint content produce estimates that read fluently but lack dimensional grounding: hallucinated quantities, confused units, and missing trade-specific logic. Flat retrieval over blueprint chunks improves factual grounding but mixes foundation measurements with roofing specifications in a single context window, destroying the trade-level coherence that human estimators maintain naturally. And the visual layer is its own obstacle: the elements a takeoff is billed from---framing members, wall runs, roof planes---must first be seen, counted, and measured across sheets drawn at different scales (Figure~\ref{fig:detail}), a perception task that frontier vision--language models do not yet perform reliably on construction documents~\cite{LinEtAl2025MeasureBench,KhangaonkarEtAl2026Spatial}. These are three different failures---missed visual structure, missing domain convention, and unmanaged context---and Section~\ref{sec:system} builds one layer for each.

We present \textbf{Handoff-H1}, a takeoff system built on the position that no single capability closes this gap. Handoff-H1 combines purpose-built computer-vision models for construction drawings; \emph{vision agents} that pair generic image operations with in-house visual-task tools; a persistent, hierarchically structured project foundation in the spirit of hierarchical retrieval-augmented generation~\cite{HuangEtAl2025}, grounded in a curated construction knowledge base. The system powers the takeoff features of the Handoff product. On the Construction Blueprint Takeoff Benchmark (\bench)---built on consensus-validated expert takeoffs whose construction we describe in full---Handoff-H1 reaches an 81.6\% composite (86.1\% material coverage, 78.8\% Precision@25\%), while seven frontier and open-weight models, run as tool-using agents from the same raw PDFs under the same evaluation harness, span composites of 35 to 61. Using the same harness, independent professional estimators were evaluated to devise a Human Performance score on all cases reaching 77.6\% composite, in comparison we see that Handoff-H1 gives up a few points of quantity precision in exchange for far higher coverage, thus creating a more thorough takeoff on items and trades that humans tend to miss.

Concretely, this report contributes: (i) a categorical description of the Handoff-H1 architecture---perception, knowledge, and orchestration---with the design rationale for each layer; (ii) a self-contained account of \bench, including the consensus process by which expert ground truth was constructed and audited; (iii) a three-way evaluation under the same per-trade LLM judge---seven frontier and open-weight models, independent professional estimators, and Handoff-H1---including a per-trade analysis of where each excels; and (iv) an access model for the benchmark---public evaluation harness, data upon request---chosen to keep its gold standard out of training corpora.

\section{Related work}
\label{sec:related}
\paragraph{Retrieval-augmented generation and hierarchical knowledge.}
RAG~\cite{LewisEtAl2020} augments generation with retrieved context, typically via dense retrieval over an embedded chunk index; the Naive/Advanced/Modular taxonomy~\cite{GaoEtAl2023RAGSurvey} places Handoff-H1 in the Modular family, with a structured synthesis layer between retrieval and generation. Hierarchical architectures organize retrieved knowledge into multi-granular structures: HiRAG~\cite{HuangEtAl2025} clusters entities into local, bridge, and global layers and attributes up to 16\% of multi-hop performance to the bridge; RAPTOR~\cite{SarthiEtAl2024RAPTOR} builds a recursive abstractive tree; GraphRAG~\cite{GraphRAG2024} and LightRAG~\cite{LightRAG2024} retrieve over entity-relation graphs. Handoff-H1's project foundation occupies a related niche, with the bridge granularity fixed by a construction-trade schema rather than learned clustering. The foundation's persistence follows the LLM-Wiki pattern~\cite{Karpathy2025}---curated knowledge maintained across queries rather than re-derived per query---applied at the retrieval index rather than the model weights, in contrast to domain-specialized pretraining such as BloombergGPT~\cite{WuEtAl2023BloombergGPT}. Domain-adapted RAG~\cite{ZhangEtAl2024RAFT} and self-reflective retrieval~\cite{AsaiEtAl2024SelfRAG} address adjacent error modes.

\paragraph{Agentic systems and per-component optimization.}
Tool-using agents that perform reasoning and acting~\cite{YaoEtAl2023ReAct} and self-refinement architectures that critique and revise their own outputs~\cite{MadaanEtAl2023SelfRefine} are standard components of production LLM pipelines; Handoff-H1 adopts both patterns to adapt in the case of conflicting information.

\paragraph{Construction-drawing datasets and the takeoff ground-truth gap.}
Public datasets for architectural drawings target recognition rather than estimation: CubiCasa5K~\cite{KalervoEtAl2019CubiCasa} provides 5{,}000 densely annotated floor plans; FloorPlanCAD~\cite{FanEtAl2021FloorPlanCAD} offers roughly 16{,}000 vector drawings for panoptic symbol spotting; ArchCAD-400K~\cite{luo2025archcad400klargescalecaddrawings} scales symbol spotting to hundreds of thousands of drawings; ResPlan~\cite{ResPlan2025} contributes 17{,}000 residential plans as vector graphs. None carries the material quantities a takeoff requires. On the estimation side, CEQuest~\cite{CEQuest2025} probes drawing literacy with 164 questions rather than full line-item takeoffs. Earlier automation splits between BIM-based takeoff, which assumes complex modeling alongside structured input~\cite{KhosakitchalertEtAl2019BIM}, and understanding of the floor-plan through deep-learning symbol detection~\cite{RezvanifarEtAl2020Symbol}. Spatial understanding from drawings remains unsolved, \citet{PeterssonEtAl2025BlueprintBench} report leading systems at or below a random baseline on floor-plan reconstruction. The scarcity of ground truth for takeoff is structural: a complete takeoff is a heavy expert lift, and estimators genuinely disagree on continuous-measurement quantities, so a defensible gold standard demands reconciliation rather than a single annotator (Section~\ref{sec:groundtruth}).

\paragraph{Specialized AI systems for takeoff vs.\ general-purpose document AI.}
Multimodal large language models have enabled a new category of AI document assistants that work directly on construction drawings, integrating OCR, natural language understanding, document retrieval, and PDF interaction. Bluebeam Max, powered by Claude~\cite{BluebeamMax2026}, is one of the first commercially available assistants embedded in a professional construction-document workflow, aimed at boosting estimator productivity through document comprehension and interactive assistance. We conducted experiments with Bluebeam Max and several other general-purpose assistants on projects covering the benchmark’s trade range, but we do not analyze them in depth because tools in this category are not intended to execute a takeoff fully autonomously, but rather to respond to user actions as the operator performs a takeoff or navigates a set of drawings. In that capacity, Bluebeam Max performed reasonably well in our tests—delivering dependable text search and extraction, interpretation of schedules and callouts, scale-aware measurements, and markup generation when steered by the user—albeit with some shortcomings. Its understanding is constrained to the document’s embedded text layer, so rasterized sheets and image-only schedules were reported as unreadable, and its page-scoped context did not reconcile quantities across the sheets of a set, leading to double-counting of doors shown on multiple sheets and missing quantities outside the active page. Importantly, when prompted to perform a full takeoff, the assistant either refused or asked the user to provide the geometry it lacked, instead of hallucinating quantities—behavior that is appropriate for a system of this type. Since such assistants do not autonomously generate the structured, per-trade line items on which the benchmark’s metrics are computed, placing them on the same leaderboard would misrepresent both their intended role and the capabilities the benchmark is meant to assess.

\paragraph{Benchmark access models.}
Open release maximizes reproducibility but exposes ground truth to training-corpus contamination, which harms the validity of benchmarks over time~\cite{SainzEtAl2023Contamination}. Several evaluation efforts therefore withhold some or all of their data: FrontierMath keeps a private holdout set~\cite{GlazerEtAl2024FrontierMath}, ARC benchmarks maintain private evaluation splits~\cite{Chollet2019ARC}, and credentialed-access corpora such as MIMIC gate data behind identity verification, thorough review and is restricted under usage agreements~\cite{JohnsonEtAl2016MIMIC}. \bench adopts the same posture: the harness is public and the data is provided upon request after review (Section~\ref{sec:access}).

\section{The Handoff-H1 system}
\label{sec:system}

\subsection{Design overview}
\label{sec:overview}

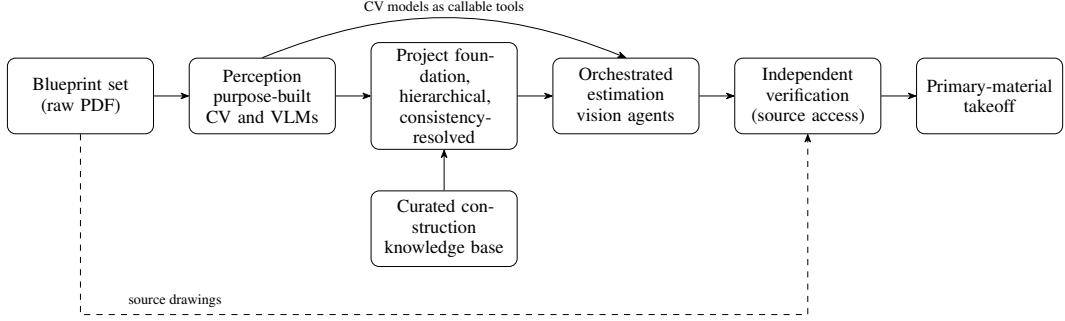
\begin{figure}[t]
  \centering
  \resizebox{\linewidth}{!}{%
  \begin{tikzpicture}[
    font=\footnotesize,>={Stealth[round]},
    box/.style={draw,rounded corners,align=center,minimum height=13mm,text width=23mm,inner sep=3pt},
    node distance=7mm and 6mm,
  ]
    \node[box] (pdf) {Blueprint set\\(raw PDF)};
    \node[box,right=of pdf] (perc) {Perception\\purpose-built CV and VLMs};
    \node[box,right=of perc] (found) {Project foundation,\\hierarchical,\\consistency-resolved};
    \node[box,right=of found] (est) {Orchestrated estimation\\vision agents};
    \node[box,right=of est] (ver) {Independent verification\\(source access)};
    \node[box,right=of ver] (out) {Primary-material\\takeoff};
    \node[box,below=of found] (kb) {Curated construction\\knowledge base};
    \draw[->] (pdf)--(perc);
    \draw[->] (perc)--(found);
    \draw[->] (found)--(est);
    \draw[->] (est)--(ver);
    \draw[->] (ver)--(out);
    \draw[->] (kb)--(found);
    \draw[->] (perc.north) to[bend left=22] node[above,font=\scriptsize]{CV models as callable tools} (est.north);
    \draw[->,dashed] (pdf.south) -- ($(pdf.south)+(0,-31mm)$) -- node[above,pos=0.13,font=\scriptsize]{source drawings} ($(ver.south)+(0,-31mm)$) -- (ver.south);
  \end{tikzpicture}%
  }
  \caption{The Handoff-H1 architecture at the level this report discusses it. Model identities, prompts, and orchestration specifics are withheld.}
  \label{fig:architecture}
\end{figure}

Handoff-H1 is organized as three layers over general-purpose base models (Figure~\ref{fig:architecture}). A \emph{perception} layer recovers the typed visual structure from the raw drawing set. A \emph{knowledge} layer---the project foundation---organizes what perception recovers, together with a curated construction knowledge base, into a persistent hierarchy. An \emph{orchestration} layer runs vision agents against the foundation and audits their output. Each layer answers a distinct failure mode of general-purpose agents on this task: missed visual structure, missing domain conventions, and unmanaged context. None of the three is the differentiator on its own; the human-level performance on the benchmark (Section~\ref{sec:results}) is the product of their combination.

\subsection{Perception: purpose-built vision models}
\label{sec:perception}

\begin{figure}[t]
  \centering
  \includegraphics[width=0.66\linewidth]{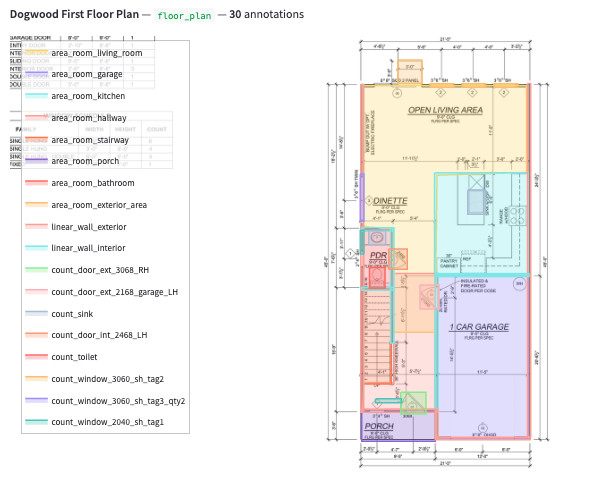}
  \caption{Reading a floor plan. The perception layer recovers typed primitives from each sheet: rooms classified by function (living, kitchen, garage, porch, bathroom), and walls, doors, windows, and fixtures detected and labeled with provenance. Discrete-count elements like these sit at the easy end of the difficulty spectrum; continuous-measurement trades (Section~\ref{sec:failures}) are substantially harder.}
  \label{fig:extraction}
\end{figure}

Before any estimate is produced, a dedicated perception pass recovers typed primitives from each sheet: rooms classified by function; walls, doors, windows, and fixtures detected and labeled; project-level dimensions and scales; and region-level structure such as plan areas and drawing callouts (Figure~\ref{fig:extraction}). This layer is built from computer-vision and vision--language models specialized on construction drawings extraction. The same models serve multiple purposes in this pipeline: they are the front end of building the knowledge layer (Section~\ref{sec:foundation}), and they are exposed as callable tools to the estimation-stage agents (Section~\ref{sec:visionagents}). Purely semantic retrieval reliably misses authoritative structural elements such as project-level dimensions or object counting; perception output is what makes those elements addressable and reliable for downstream tasks.

\subsection{Vision agents}
\label{sec:visionagents}

The estimation stage is carried out by \emph{vision agents}: tool-using agents~\cite{YaoEtAl2023ReAct} whose tool set spans two tiers. The first tier is generic image manipulation---cropping a region of a sheet, zooming to read a dimension string or a callout, panning across sheets---the operations any capable multimodal agent can be given. The second tier is task-specific: in-house tools that decompose a framing plan into its individual members, and tools that invoke the purpose-built computer-vision models of Section~\ref{sec:perception} to count, detect, and measure elements within a selected region. The agent decides when to trust its own reading of a drawing and when to delegate to a specialized tool; the second tier gives it capabilities---member-level framing decomposition, reliable counting over dense symbol fields---that prompting a general vision--language model does not provide.

\subsection{Project foundation}
\label{sec:foundation}

The project foundation is a persistent, hierarchically structured knowledge representation built once per project and consumed on demand by everything downstream. Construction proceeds in three phases. \emph{Extraction} performs dense semantic retrieval over the perception output, surfacing dimensional measurements, specification text, and visual structure with explicit provenance and authority tags. \emph{Synthesis} partitions extracted content by construction clusters and grounds each partition against the curated knowledge base; the partition plays the role of the bridge layer in the hierarchical-RAG sense~\cite{HuangEtAl2025}---a domain-meaningful intermediate granularity at which retrieval and reasoning proceed without cross-cluster context contamination. \emph{Consistency resolution} responsible for finding conflicts within partitions and solving them.

The curated knowledge base that grounds synthesis follows the LLM-Wiki pattern~\cite{Karpathy2025}: construction reference materials, estimating guides, material catalogs, and pricing data curated by domain experts into structured, searchable entries. The curation encodes logic that experienced estimators carry implicitly and it is continuously built and improved with information such as assembly conventions, default specifications, code-driven defaults---that no amount of open-domain nor in-context retrieval supplies. Its contents, sources, and curation methodology remain proprietary. Once constructed, the foundation is persisted: estimation runs consume it without re-processing the blueprint.

\subsection{Orchestration and verification}
\label{sec:orchestration}

Each vision agent works within its given scope with foundation queries and visual tools, rather than one agent attempting the whole project in a single context. An independent verification pass then audits the assembled takeoff with direct access to the source drawings---a self-refinement architecture~\cite{MadaanEtAl2023SelfRefine} adapted to retrieval-grounded outputs---catching missed items and implausible quantities before the takeoff is finalized. The output is a structured set of primary-material line items: trade section, material description, quantity, and unit of measure. Stage decomposition, routing, and verification logic are not disclosed.

\section{The Construction Blueprint Takeoff Benchmark}
\label{sec:benchmark}

Handoff-H1 is evaluated on \bench, a benchmark we built for exactly this task and describe here in full: the dataset, the consensus process behind its ground truth, the scoring protocol, and the access model.

\subsection{Dataset}
\label{sec:dataset}

\bench comprises 10 residential blueprint sets drawn from real projects that Handoff is permissioned to use. Each set is a complete drawing package---floor plans, foundation plans, elevations, wall sections, and roof plans---representative of the segment that constitutes the majority of estimation volume. Every set is stripped of personally identifying information (owner and firm names, addresses, professional stamps, and project identifiers) by an automated detection pass followed by human review. The ground truth is a clean, line-item schema: trade section, material description, quantity, and unit of measure. Every set is evaluated across nine trades: Concrete, Framing, Roofing, Siding, Insulation, Drywall, Electrical, Plumbing, and Windows \& Doors.

The ground truth is deliberately limited to \emph{primary} materials---the framing members, sheets, and structural quantities that dominate cost---and excludes secondary items (minor fasteners, incidental accessories) and labor lines. This is a substantive design choice, not an oversight: secondary quantities are largely preference-driven and not worth reconciling to a single number, so including them would inject noise into both the ground truth and the scoring. Restricting the benchmark to primary materials keeps the gold standard verifiable and the comparison fair.

Ten sets understate the evaluation's size. Each drawing package decomposes into hundreds of images and accompanying text, and evaluation is item-level: the reconciled ground truth across the benchmark carries 2{,}009 verified line items, 1{,}348 of them primary-tier and in scope for scoring. A drawing set is therefore not one task but hundreds---every primary item is its own find, measure, and specify problem, judged individually.

\subsection{Ground truth and human performance}
\label{sec:groundtruth}

A takeoff benchmark is only as trustworthy as its gold standard, and a single estimator's opinion is not a defensible one. Each ground-truth takeoff is therefore a reconciled, multi-reviewer artifact. Independent professional estimators, each with at least five years of experience, were engaged and paid to produce complete takeoffs from each drawing set according to a fixed instruction set. An internal team of construction experts then reviewed each submission in two stages. They first audited \emph{structure}---that the trades, scopes, naming conventions, and overall organization matched the benchmark's guidance; agreement at this level defines what we call human performance for the task. They then audited \emph{coverage and precision}, making corrections by hand, without AI assistance. A takeoff accepted by at least two construction experts and at least two independent estimators becomes the \emph{inter-estimator-approved} takeoff, and that artifact is the gold standard for each drawing set. The independent submissions --- with naming and structure convention ---, scored by the benchmark's own judge system against the \emph{inter-estimator-approved} takeoff, place human performance at 65.5\% coverage and 87.9\% Precision@25\%---a realistic reference that sits below 100\% on the continuous-measurement trades where estimators legitimately differ, and that Section~\ref{sec:humancomparison} compares against Handoff-H1 directly.

\subsection{Metrics}
\label{sec:metrics}

Predictions and ground truth are grouped by trade, and an LLM judge (OpenAI gpt-5.5)---now a standard, low-cost alternative to exact-match scoring for structured outputs---matches predicted items to ground-truth items within each trade, allowing many-to-many groupings so that several predicted rows can be reconciled against one ground-truth assembly. Only material line items participate; labor and non-material rows are excluded. Judging is stochastic, so no reported number rests on a single sample: every prediction set is judged ten times and the reported metrics are averages across judge runs. From the matches we compute two metrics. \textbf{Coverage} is the share of ground-truth primary items for which the model produced a matching prediction; it measures completeness. An uncovered item is not necessarily an outright omission: the judge matches on construction specification, so a prediction that names the wrong spec leaves the ground-truth line uncovered---\nicefrac{5}{8}$''$ drywall does not match a \nicefrac{1}{2}$''$ ground-truth item, and plain \nicefrac{1}{2}$''$ does not match the moisture-resistant board a bathroom calls for. \textbf{Quantity Precision@25\%} (P@.25) is the share of matched items whose predicted quantity falls within 25\% of the approved quantity. A single \textbf{composite} combines them,
\begin{equation}
  \text{score} = \text{coverage}^{0.4} \cdot \text{precision}^{0.6}.
  \label{eq:composite}
\end{equation}
Precision and coverage are combined multiplicatively so that the composite score rewards estimates that are both comprehensive and accurate. Precision alone is trivially gamed: a model that reports a single correct line item and omits the remainder of the takeoff would post 100\%. Scaling by coverage removes this degenerate case, since a score can only be high when most of the takeoff is recovered and the recovered items are correct.

The exponents weight precision more heavily than coverage. This asymmetry is deliberate. The ground truth is more reliable about the items it contains than about the items it omits, an uncovered item may reflect a gap in the gold standard rather than a genuine omission, whereas a wrong quantity on a matched item is an unambiguous failure. Coverage misses are therefore penalized more leniently than precision errors.

Precision is scored against a 25\% tolerance rather than a tighter band. Differences within 10\% are common even when a single professional remeasures their own prior work, particularly for items quantified by linear foot or square foot, where ordinary measurement variance can account for a discrepancy of that magnitude. A 10\% threshold would consequently penalize measurement noise rather than genuine error.

The four measurement types---count, linear, area, and volumetric---are collapsed into a single score rather than reported individually. This simplifies comparison across runs, at the cost of overpenalizing counts: an off-by-one error typically exceeds the 25\% tolerance, so counting rarely falls within the band that dimensional measurements comfortably occupy.

\subsection{Evaluation protocol}
\label{sec:conditions}

Every system starts from the same input: the blueprint PDF and nothing else. Each baseline model runs as an agent that must emit a structured set of line items per trade. Because some language models are not natively multimodal, the harness provides a multimodality equalizer: for those models, regions of interest are supplied as text descriptions together with a visual-analysis tool (backed by Gemini~3.1~Pro) that answers targeted visual questions about a region on demand, so the comparison turns on reasoning rather than image support. Baseline agents are run up to three times per blueprint set and their scores averaged. Handoff-H1 takes no assistance from the harness: it runs end-to-end from the raw PDF, since the perception layer is part of the system being measured. The independent estimators' takeoffs (Section~\ref{sec:groundtruth}) are scored by the identical judge against the same approved ground truth, which is what makes the three-way comparison of Section~\ref{sec:results} comparable.

\subsection{Availability and access}
\label{sec:access}

The evaluation harness is public. \bench is packaged for the open \texttt{harbor} framework under Apache-2.0: each task is a \texttt{(blueprint, trade)} pair in which the agent writes a predictions file and a separate verifier runs the judge against the held-out ground truth and emits a per-task reward (Appendix~\ref{app:harbor}).

The blueprint sets and ground-truth takeoffs are available upon request. Researchers and practitioners can request access through the benchmark repository\footnote{https://github.com/handoffai/residential-takeoff-benchmark} or by writing to \href{mailto:research@handoff.ai}{research@handoff.ai}; requests are reviewed before the data is provided. We adopt this access model for one reason above others: a benchmark whose gold standard circulates in web crawls eventually measures memorization rather than capability~\cite{SainzEtAl2023Contamination}, and expert-consensus takeoff ground truth is too expensive to burn that way. Withholding evaluation data behind a request step is by now standard practice~\cite{GlazerEtAl2024FrontierMath,Chollet2019ARC,JohnsonEtAl2016MIMIC}; it also lets us honor the terms under which the underlying projects were permissioned. Leaderboard submissions remain open---request the data, run the public harness, and submit results---and we welcome research partnerships and individual researchers interested in the problem.

\paragraph{What the evaluation captures---and what it does not.}
Two caveats bound the interpretation of \bench scores. First, matching is performed by an LLM judge: it tolerates differences in labeling conventions that exact-match scoring would miss, but it is not a substitute for a human adjudicator on genuinely ambiguous rows---and averaging ten judge runs controls sampling variance, not any systematic bias the judge may carry. Second, Precision@25\% is measured against a single reconciled quantity, yet estimators themselves disagree on continuous-measurement trades---whether a dimension string includes adjacent wall thickness, how a sloped roof reconciles against its plan footprint---so the achievable ceiling there is below 100\% even for careful humans (Section~\ref{sec:groundtruth}).

\section{Results}
\label{sec:results}

This section reports the complete evaluation over the 10 benchmark sets: the main comparison against seven frontier and open-weight models (Section~\ref{sec:leaderboard}), the comparison against independent professional estimators (Section~\ref{sec:humancomparison}), and the per-trade analysis of where each excels (Section~\ref{sec:pertrade}).

\subsection{Main comparison}
\label{sec:leaderboard}

\begin{figure}[t]
  \centering
  \includegraphics[width=0.92\linewidth]{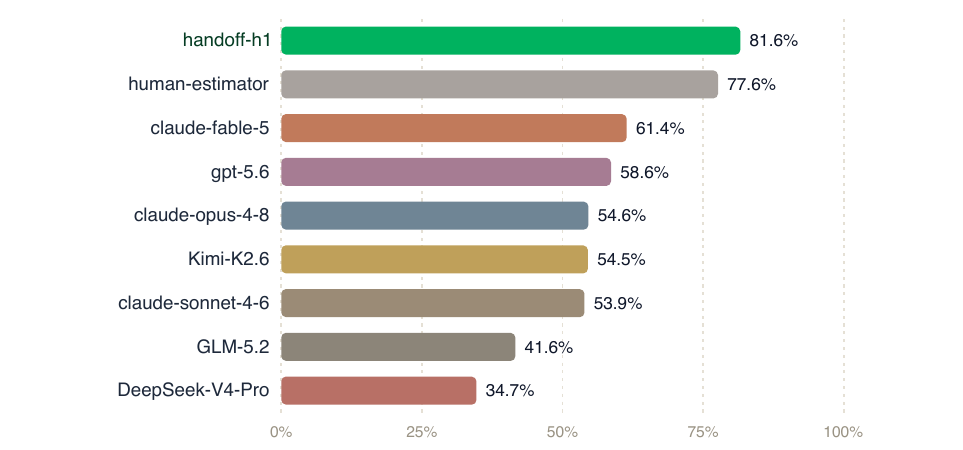}\\[8pt]
  \includegraphics[width=0.98\linewidth]{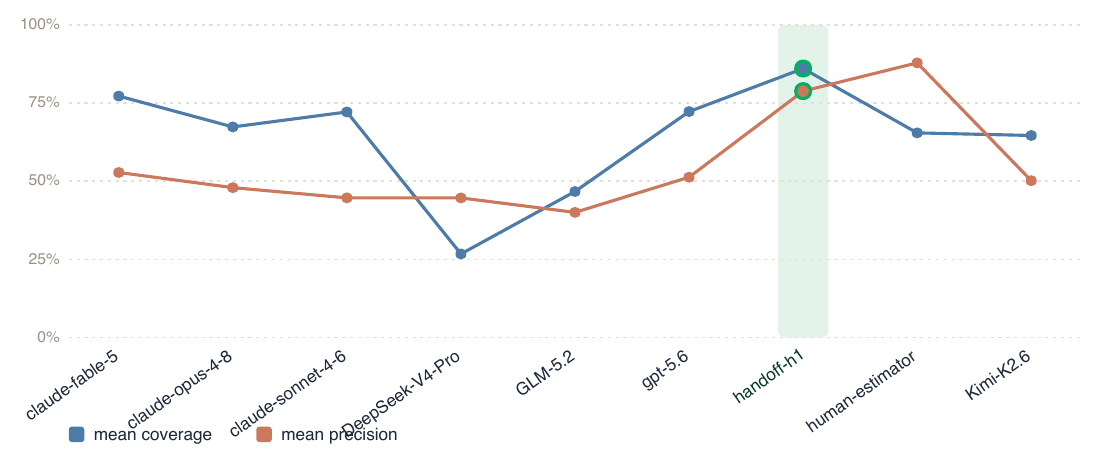}
  \caption{Top: composite score by system (Equation~\ref{eq:composite}), Handoff-H1 highlighted. Bottom: the mean coverage and precision beneath each composite, systems in alphabetical order.}
  \label{fig:cov-acc}
\end{figure}

Table~\ref{tab:leaderboard} reports the main comparison: seven frontier and open-weight models, each run as a tool-using agent from the raw PDF, scored under the identical per-trade judge against Handoff-H1 and against the independent professional estimators of Section~\ref{sec:groundtruth}. Figure~\ref{fig:cov-acc} shows the composite ranking and the coverage and precision underlying it.

\begin{table}[t]
  \caption{Main comparison on \bench (all 10 sets). Every system works from the raw blueprint PDF; baselines additionally receive the visual-analysis equalizer where needed, and their scores are averaged over repeated runs. The independent estimators' takeoffs are scored by the same judge against the same approved ground truth. Composite is defined by Equation~\ref{eq:composite}; values are percentages, higher is better.}
  \label{tab:leaderboard}
  \centering
  \begin{tabular}{lccc}
    \toprule
    System & Composite & Coverage & P@.25 \\
    \midrule
    \textbf{Handoff-H1} \textit{(end-to-end)} & \textbf{81.6} & \textbf{86.1} & 78.8 \\
    Human estimators \textit{(independent)}   & 77.6 & 65.5 & \textbf{87.9} \\
    \midrule
    claude-fable-5 \textit{(closed)}          & 61.4 & 77.3 & 52.8 \\
    gpt-5.6 \textit{(closed)}                 & 58.6 & 72.3 & 51.3 \\
    claude-opus-4-8 \textit{(closed)}         & 54.6 & 67.4 & 48.0 \\
    Kimi-K2.6 \textit{(open)}                 & 54.5 & 64.6 & 50.2 \\
    claude-sonnet-4-6 \textit{(closed)}       & 53.9 & 72.2 & 44.7 \\
    GLM-5.2 \textit{(open)}                   & 41.6 & 46.7 & 40.0 \\
    DeepSeek-V4-Pro \textit{(open)}           & 34.7 & 26.7 & 44.7 \\
    \bottomrule
  \end{tabular}
\end{table}

\paragraph{No frontier agent reaches the estimators.} The seven models span composites of 34.7 to 61.4, all below the independent-estimator reference of 77.6. The strongest is claude-fable-5 (61.4), an Anthropic Mythos-class model positioned above the Opus tier~\cite{AnthropicFable5_2026}; gpt-5.6 follows at 58.6. The closed models hold the top of the field, but the edge is not categorical: the strongest open-weight model (Kimi-K2.6, 54.5) is level with claude-opus-4-8 (54.6). Base-model choice moves a takeoff agent by tens of points---and still leaves the best of them sixteen short of a professional estimator.

\paragraph{Handoff-H1 opens a 20-point gap from identical inputs.} Handoff-H1 reaches 81.6 composite, roughly 20 points above the best frontier agent, from exactly the same starting point: the raw PDF. The gap is not bought with a stronger base model---the frontier field's internal spread shows how little that buys on this task; it is the perception, knowledge, and orchestration layers that the baselines do not have (Section~\ref{sec:system}).

\paragraph{Frontier agents see more than they can measure.} The failure profile of the frontier field is one-sided (Figure~\ref{fig:cov-acc}): coverage exceeds precision for every one of the seven models---claude-fable-5 finds 77.3\% of the primary items but quantifies only 52.8\% of its matches within the 25\% band. The bottleneck for a capable general agent is no longer locating the scope of a takeoff; it is the dimensional reasoning that turns a found item into a defensible quantity. The human estimators fail in exactly the opposite direction (65.5 coverage, 87.9 precision), which sets up the comparison of Section~\ref{sec:humancomparison}.

\subsection{Comparison with professional estimators}
\label{sec:humancomparison}

The practical baseline for a takeoff system is not a language model; it is a professional estimator. We therefore compare Handoff-H1 with the held-out independent expert takeoff for each drawing set (Section~\ref{sec:groundtruth}), both scored by the same judge against the same approved ground truth.

Handoff-H1 posts the higher composite on six of the ten sets, the estimators on four (one of them a half-point wash). The mechanics behind the split are consistent. \emph{The estimators' losses are coverage.} Their takeoffs often missed primary items that impacted coverage largely on Structural Framing members - such as joists, rafters and beams - and items that are only called-out on drawing details such as drywall sheets of project-specified thickness and rebar quantities, which require reading complex details and combining details with foundation plans. Handoff-H1 had the edge on this front largely due to completeness self-revision when performing takeoffs (Section~\ref{sec:orchestration}). \emph{The estimators win on precision.} On the sets they win, they measure what they take off at 89\% P@.25 or better, and the single best takeoff in the pool is human work. But the two differ in consistency as much as in average score: Handoff-H1's per-set composite stays within a nine-point range, while the estimators' swings more than thirty. Humans produce both the best takeoffs and the worst; the system gives up the high peak in exchange for a floor the human pool never reaches. For takeoffs that feed downstream estimates, that steadiness matters more than an occasional high score.

\subsection{Per-trade quality}
\label{sec:pertrade}

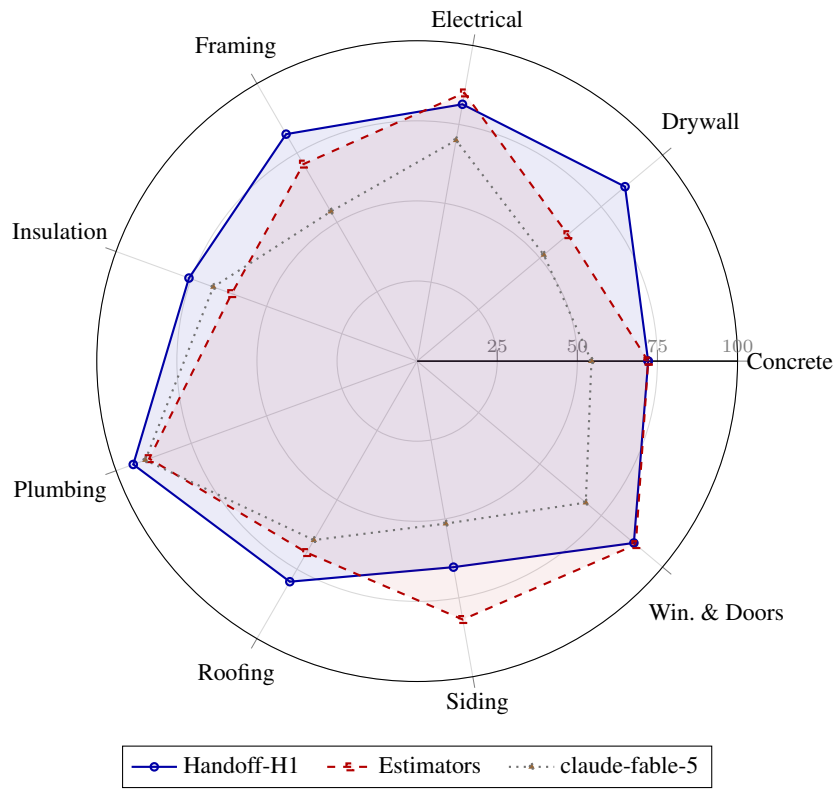
\begin{figure}[t]
  \centering
  \begin{tikzpicture}
    \begin{polaraxis}[
      width=0.72\linewidth,
      xmin=0, xmax=360,
      ymin=0, ymax=100,
      xtick={0,40,80,120,160,200,240,280,320},
      xticklabels={Concrete,Drywall,Electrical,Framing,Insulation,Plumbing,Roofing,Siding,Win.\ \& Doors},
      xticklabel style={font=\small},
      ytick={25,50,75,100},
      yticklabel style={font=\scriptsize,gray},
      grid=both, grid style={gray!30},
      legend style={at={(0.5,-0.10)},anchor=north,legend columns=3,font=\small,/tikz/every even column/.append style={column sep=8pt}},
    ]
      \addplot+[color=blue!65!black,mark=*,mark size=1.4pt,thick,fill=blue!65!black,fill opacity=0.08]
        coordinates {(0,72.1) (40,84.7) (80,81.4) (120,81.8) (160,75.8) (200,94.3) (240,79.5) (280,65.3) (320,88.3) (360,72.1)};
      \addlegendentry{Handoff-H1}
      \addplot+[color=red!70!black,mark=square*,mark size=1.3pt,thick,dashed,fill=red!70!black,fill opacity=0.06]
        coordinates {(0,72.1) (40,61.4) (80,85.0) (120,71.0) (160,61.6) (200,89.0) (240,69.0) (280,81.9) (320,89.1) (360,72.1)};
      \addlegendentry{Estimators}
      \addplot+[color=black!55,mark=triangle*,mark size=1.4pt,thick,dotted]
        coordinates {(0,54.4) (40,51.6) (80,70.0) (120,53.9) (160,67.7) (200,90.3) (240,64.5) (280,51.5) (320,68.7) (360,54.4)};
      \addlegendentry{claude-fable-5}
    \end{polaraxis}
  \end{tikzpicture}
  \caption{Per-trade composite as a radar: Handoff-H1 (solid), the independent estimators (dashed), and the strongest frontier agent, claude-fable-5 (dotted). The estimator profile collapses where scopes went unquantified (Drywall, Insulation) and dominates on Siding; the frontier agent sits inside both profiles on seven of the nine trades.}
  \label{fig:radar}
\end{figure}

Table~\ref{tab:pertrade} breaks the composite down by trade for all nine systems, and Figure~\ref{fig:radar} draws the three-way comparison---Handoff-H1, the independent estimators, and the strongest frontier agent---as a radar over the nine trades.

\begin{table}[t]
  \caption{Per-trade composite score on \bench, pooled over all 10 sets (baselines also over repeated runs). Values are percentages; bold marks the per-trade leader. Handoff-H1 leads five of the nine trades; the independent estimators lead on Siding decisively and on Electrical and Windows \& Doors by small margins; Concrete is tied; no frontier agent leads any trade.}
  \label{tab:pertrade}
  \centering
  \resizebox{\linewidth}{!}{%
  \begin{tabular}{lccccccccc}
    \toprule
    Trade & H1 & Estimators & fable-5 & gpt-5.6 & opus-4-8 & Kimi-K2.6 & sonnet-4-6 & GLM-5.2 & DeepSeek-V4 \\
    \midrule
    Concrete         & \textbf{72} & \textbf{72} & 54 & 51 & 45 & 43 & 46 & 30 & 25 \\
    Framing          & \textbf{82} & 71 & 54 & 49 & 38 & 53 & 39 & 31 & 29 \\
    Roofing          & \textbf{79} & 69 & 64 & 60 & 62 & 55 & 56 & 34 & 32 \\
    Siding           & 65 & \textbf{82} & 51 & 48 & 42 & 50 & 49 & 38 & 26 \\
    Insulation       & \textbf{76} & 62 & 68 & 65 & 53 & 51 & 46 & 44 & 33 \\
    Drywall          & \textbf{85} & 61 & 52 & 54 & 56 & 57 & 45 & 33 & 29 \\
    Electrical       & 81 & \textbf{85} & 70 & 69 & 68 & 68 & 72 & 65 & 57 \\
    Plumbing         & \textbf{94} & 89 & 90 & 89 & 89 & 70 & 82 & 63 & 42 \\
    Windows \& Doors & 88 & \textbf{89} & 69 & 66 & 68 & 58 & 66 & 57 & 47 \\
    \bottomrule
  \end{tabular}%
  }
\end{table}

Three patterns organize the field. First, \emph{counting-like trades are where everyone converges}: Plumbing is the best or near-best trade for every system in the table---fixtures are enumerable---and the margins there are small (Handoff-H1 94, estimators 89, claude-fable-5 90); Windows \& Doors behaves the same way, with the estimators (89) a point ahead of H1 (88). Second, \emph{derivation-heavy trades are where Handoff-H1 pulls away from both references}: its four largest margins over the estimators are Drywall ($+23$), Insulation ($+14$), Framing ($+11$, the trade its member-level decomposition tooling of Section~\ref{sec:visionagents} targets directly), and Roofing ($+10$)---exactly the trades whose quantities and item specifications must be derived from areas, pitches and multi-sheet specification reasoning rather than read off single plans, and exactly where the estimators' skipped scopes concentrate. The frontier agents bottom out on the same trades (no model above 57 on Drywall or 54 on Framing), so the derivation problem is not solved by scale. Third, \emph{the estimators keep genuine territory}: Siding is their decisive win (82 against 65, Handoff-H1's weakest trade), they edge the system on Electrical (85 against 81) and Windows \& Doors, and with Concrete tied between H1 and estimators. Among the frontier agents the per-trade ranking reshuffles---Kimi-K2.6 overtakes both Claude models on Framing and Drywall, claude-sonnet-4-6 leads the models on Electrical---evidence that they fail in different places; with gpt-5.6 and fable-5 standing standouts primarily on precision amongst the other models tested.

\section{Failure modes and limitations}
\label{sec:failures}

\paragraph{Precision trails coverage.} At 86.1\% coverage and 78.8\% P@.25, the system's remaining errors concentrate on quantities rather than omissions or spec-level errors: more than one matched quantity in five still falls outside the 25\% band, where the human estimators hold 87.9\%. The asymmetry is structural. Coverage failures are discrete---an item omitted outright, or taken off under the wrong specification and left unmatched (Section~\ref{sec:metrics})---and respond to orchestration, verification, and knowledge grounding; precision depends on dimensional reasoning, unit conversions, and trade conventions that are intrinsically harder to fix from the estimation stage alone, this is why improvements to the curated knowledge base and to the vision components are a continuous task.

\begin{figure}[t]
  \centering
  \includegraphics[width=0.7\linewidth]{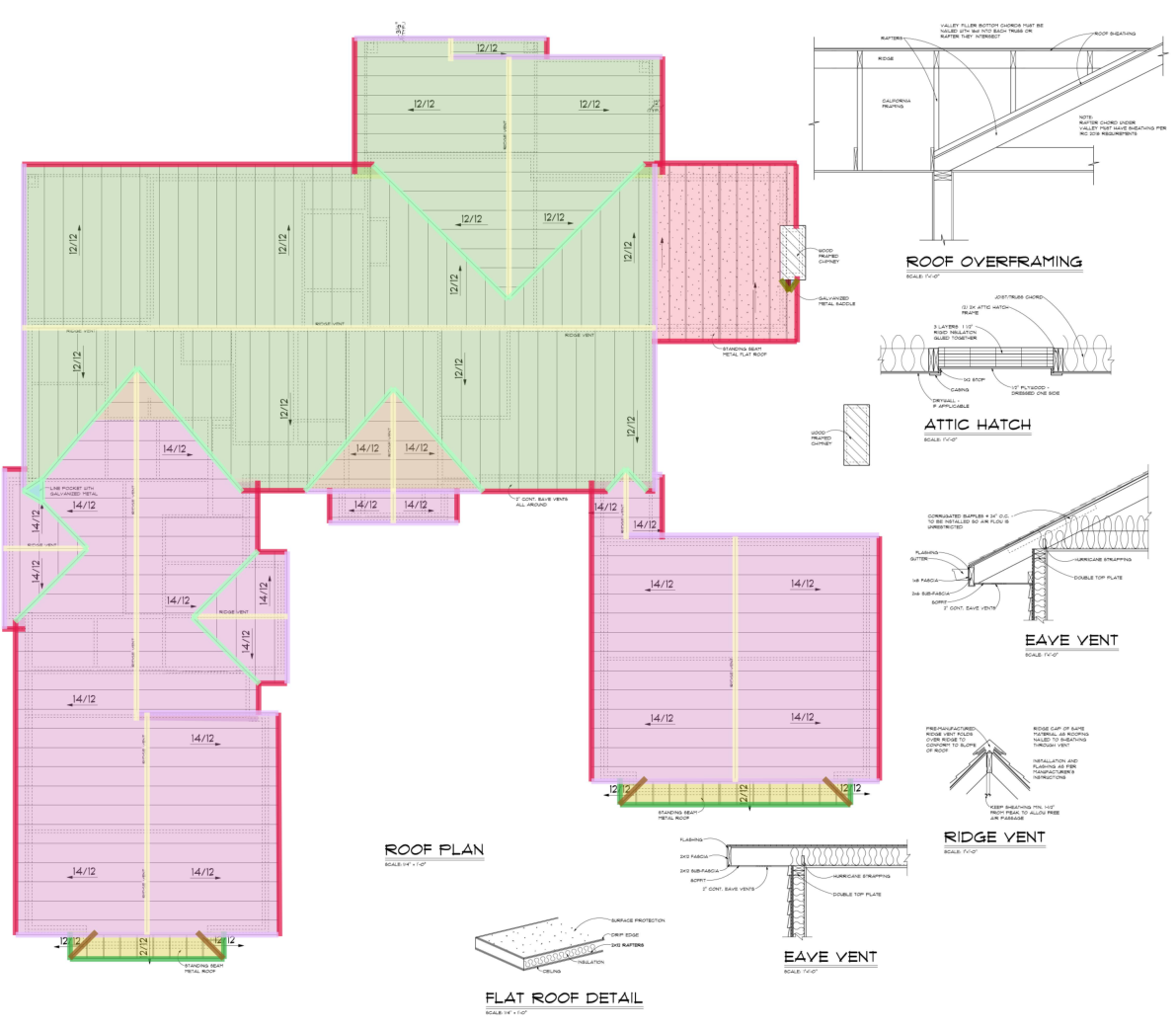}
  \caption{A roof plan with surfaces broken out by plane and pitch. Roofing is sold in squares, and the true area depends on a slope the plan view hides: the billed quantity must be recovered from pitch and framing information rather than read off the flat footprint. Continuous-measurement trades like this are where estimators legitimately disagree, and where both Handoff-H1 and the human ceiling fall furthest from a single gold number.}
  \label{fig:roof}
\end{figure}

\paragraph{Continuous-measurement trades remain the hard core.} Trades measured in linear or square feet (framing, siding, roofing) show consistently higher error rates than discrete-count trades (doors, windows, fixtures), in Handoff-H1 and in every baseline (Table~\ref{tab:pertrade}). Wall lengths must be inferred from floor-plan dimensions, requiring resolution of ambiguities---whether a dimension line includes or excludes adjacent wall thickness---that even human estimators debate. Roofing compounds the problem: the area billed is the sloped area, which the plan view foreshortens, so the quantity must be recovered from pitch and framing information rather than read off the footprint (Figure~\ref{fig:roof}). These are the trades where the human reference itself sits furthest below 100\% (Section~\ref{sec:groundtruth})---and Siding, the one trade where the estimators beat Handoff-H1 decisively (Section~\ref{sec:pertrade}), is the system's clearest open problem in this family due to the possible variations in drawing convention - where materials do not follow a clear visual convention thus making it harder to treat it as a computer vision task.

\paragraph{Knowledge-base maintenance.} The curated construction knowledge base requires ongoing maintenance as building codes evolve, material standards change, and pricing data ages. Regional variation in construction practice is a particular challenge.

\paragraph{Evaluation scope.} Results are reported on residential blueprints. Commercial construction, multi-story buildings, and projects with partial or inconsistent drawing sets may behave differently and remain to be evaluated.

\section{Discussion and next steps}
\label{sec:discussion}

\paragraph{Human comparison.} The practical baseline for this task is a professional estimator, and Section~\ref{sec:humancomparison} now measures against one directly: on this benchmark Handoff-H1 reaches human-level performance and surpasses it on several fronts---the composite itself (81.6 against 77.6 on average), six of the ten sets, and five of the nine trades. The overall quality of the takeoff is higher because the system covers more ground, and covering more ground is the harder road: every additional item quantified is another item scored, so Handoff-H1's 78.8\% precision is earned over roughly a third more matched items than the estimators' 87.9\%, which is computed only over the items and specs that are accurately detected and named.

\paragraph{The gap against frontier models.} Every baseline starts from exactly what Handoff-H1 starts from---the raw PDF---and the 20-point gap to the strongest frontier agent persists. This locates the difference not in any single ingredient but in the combination the system is built around: perception the agents can call on demand, knowledge partitioned the way estimators actually work, and orchestration that keeps each trade's reasoning scoped and verified.

\paragraph{Next steps.} We will keep the comparison current as new frontier models ship, extend evaluation beyond the residential segment, and broaden the estimator pool so the human reference reflects more than one professional takeoff per set. Requests for benchmark access, leaderboard submissions, research partnerships, and individual collaborations are welcome at \href{mailto:research@handoff.ai}{research@handoff.ai}.

\section{Conclusion}
\label{sec:conclusion}

We presented Handoff-H1, a generally available system for material quantity takeoff from construction blueprints, and the benchmark we built to measure it. The system's design commitment is combination over silver bullet: purpose-built vision models, vision agents with task-specific visual tools, a persistent partitioned project foundation grounded in a curated knowledge base, and orchestrated estimation with independent verification. On \bench---10 permissioned residential drawing sets with consensus-validated, primary-only expert takeoffs---Handoff-H1 reaches an 81.6\% composite (86.1\% coverage, 78.8\% P@.25) from the raw PDF, roughly 20 points above the strongest of seven frontier and open-weight agents run from the same input, and above the independent professional estimators scored against the same gold standard (77.6\%), whose takeoffs are often more precise but far less thorough. The evaluation harness is public; the data is available upon request, an access model chosen so that the benchmark keeps measuring capability rather than memorization. We invite the community to request access, submit results, and pressure-test the gap.

\appendix

\section{Running the benchmark with \texttt{harbor}}
\label{app:harbor}

The evaluation harness runs on the open \texttt{harbor} framework and is public under Apache-2.0; the blueprint sets and ground truth are delivered upon an approved access request (Section~\ref{sec:access}). Each task is a single \texttt{(blueprint, trade)} pair: an agent emits predicted line items and a verifier grades them against the held-out ground truth.

\begin{verbatim}
# clone the public harness
git clone https://github.com/handoffai/residential-takeoff-benchmark

# place the requested data under tasks/ once access is granted

# run one (blueprint, trade) task on the raw-PDF track
harbor run -p tasks/<case>-<trade> -a handoff -m <provider>/<model>

# aggregate per-task rewards into the leaderboard
harbor report
\end{verbatim}

\bibliography{biblio}

\end{document}